\documentclass{article}

\usepackage[dblblindworkshop,final]{neurips_2026}
\workshoptitle{Trustworthy AI for Good (AI4GOOD)}

\usepackage{newtxtext,newtxmath}
\usepackage{graphicx}
\usepackage{booktabs}
\usepackage{amsmath}
\usepackage{array}
\usepackage{microtype}
\usepackage{xurl}
\usepackage{xcolor}
\usepackage{tikz}
\usetikzlibrary{arrows.meta,positioning,fit,backgrounds,calc}
\usepackage[colorlinks=true,allcolors=blue]{hyperref}
\hypersetup{pdftitle={Tracing Whose Voices Survive AI Summarization: An Evidence-Linked Audit Interface for Civic Synthesis},pdfauthor={Sachit Mahajan}}

\definecolor{ink}{HTML}{1F2937}
\definecolor{muted}{HTML}{64748B}
\definecolor{accent}{HTML}{2B6CB0}
\definecolor{surface}{HTML}{F1F5F9}
\definecolor{panelfill}{HTML}{EDF3FA}
\newcolumntype{Y}[1]{>{\raggedright\arraybackslash}p{#1}}

\title{Tracing Whose Voices Survive AI Summarization: An Evidence-Linked Audit Interface for Civic Synthesis}

\author{Sachit Mahajan \\
  Computational Social Science, ETH Zurich \\
  \texttt{sachit.mahajan@gess.ethz.ch}}

\begin{document}
\maketitle

\begin{abstract}
Language-model summaries can make a contested civic record appear more settled than the underlying contributions. Qualifications, minority positions, and unresolved disagreements may disappear even when a summary remains fluent and factually plausible. This paper presents Provenance Lab, an open-source application for tracing summary sentences and claims to candidate source passages. The interface keeps weak, missing, and unassessed cases visible and carries the evidence through review, revision, and export. In two meeting benchmarks, a semantic search model ranked relevant transcript passages higher on average than keyword search did. The same model did a better job of matching policy arguments to expert-written key points. ExplainMeetSum, one of the meeting benchmarks, includes human annotations that identify passages supporting the reference summaries. The model ranked one of those passages first 55.9\% of the time, on average across the 35 meetings in that benchmark. Provenance Lab therefore shows several passages so reviewers can look beyond the first match. In published deliberation data, similarity to an individual's contribution modestly improved prediction of that participant's preferences. Adding minimum alignment and the Gini coefficient together yielded no detectable improvement. These results support the use of retrieval to locate candidate evidence. Whether the interface helps facilitators find more omissions or make better revisions remains an open question.
\end{abstract}

\section{Introduction}

AI-assisted synthesis is entering public consultation, meeting documentation, survey analysis, and participatory research. In these settings a summary can become the version of the record that officials, facilitators, and participants actually encounter. It shapes which problems appear shared, which reasons appear marginal, and which proposed conditions remain attached to a recommendation. Fluency and factual consistency matter, yet neither reveals whose contribution remains legible after many voices have been compressed into one account.

Participation is more than topic presence. Contributors can discuss the same policy while advancing different reasons, experiencing different consequences, or supporting action under incompatible conditions. A summary may retain the topic and remove the distinctions \citep{zhang2024,vanDerMeer2024}. This paper calls that loss \emph{representational erasure}: source meaning becomes difficult to recover from the synthesized account. The term describes an observable relation between source and summary and makes no claim about the author's intention. A related risk is \emph{apparent consensus}, in which the summary expresses convergence more strongly than the source record warrants. That appearance may be deliberately engineered, or it may arise from length constraints, majority patterns, prompting, and the tendency of summarization systems to produce coherent prose. In either case, the governance problem is the same. Readers need a way to inspect the evidence behind the collective voice.

Research on citizen participation and deliberation has long treated voice, reason giving, contestation, and the conditions of inclusion as central concerns \citep{arnstein1969,fishkin2018,landemore2020}. Participatory AI and pluralistic alignment bring these concerns into computational design \citep{delgado2023,mun2024,sorensen2024,kirk2024}. Data provenance records where an artifact came from and which activities produced it \citep{buneman2001,simmhan2005,moreau2013}. Attribution and factuality research defines source support, identifies local evidence, or verifies atomic claims \citep{rashkin2023,slobodkin2024,min2023}, while opinion summarization examines coverage across perspectives \citep{hosking2023,zhang2024,vanDerMeer2024}. Documentation frameworks describe data and models \citep{gebru2021,mitchell2019}. These traditions supply important pieces, but they leave a practical gap between a collective summary and the people or records it claims to represent.

Provenance Lab is an open-source tool for auditing AI-assisted civic summaries. The audit starts with the original contributions and asks whose views, reasons, and conditions remain traceable in the summary. A reviewer can select a submission, open its closest matches in the summary, and compare the wording to check for omissions or changes in meaning. The reviewer can also follow a summary claim back to candidate passages in the source record. Aggregate indicators help reviewers find cases to inspect in the original text.

This paper builds on the concept of participatory provenance, introduced in earlier work on public consultation \citep{mahajan2026participatory}. Provenance Lab implements a framework for auditing how summaries represent their inputs. Reviewers can inspect flagged passages, record challenges, and compare revisions against the same source contributions. The tool keeps the record of how the summary was produced separate from the similarity links it computes. It marks missing and unassessed cases and withholds analyses when the required evidence is unavailable. Keeping these limits visible reduces the risk that a dashboard lends undue authority to an uncertain summary. The evaluation tests evidence retrieval against three external reference datasets.

The intended role is review and contestation before a summary acquires institutional authority. Provenance Lab supplies no fairness score and issues no certificate. It begins with the records provided to it, so it cannot reveal people who were excluded before data collection. Semantic association cannot establish causal influence, agreement, or democratic legitimacy. These boundaries are part of the contribution because an audit interface should narrow uncertainty without converting computational evidence into political authorization.

\section{Representation as an auditable process}

Three requirements translate the representational problem into an interface.

\textbf{Distinguish process history from representational evidence.} A record that a model produced a summary is an observed event. A similarity link between a contribution and a sentence is an inferred association. The latter can guide inspection, but it cannot show that the contribution caused the sentence or that both express the same position. Provenance Lab preserves this distinction in the interface and in its portable record, where semantic links appear as assessments rather than derivations.

\textbf{Treat absence as information.} Representational erasure can be compounded by measurement failure. A record may have no credible summary match, or an automated stage may fail to assess it. Collapsing both into a low score would hide the difference. Every rate therefore travels with its eligible denominator, assessment coverage, and abstentions. Missing model output remains visible instead of silently disappearing from the audit.

\textbf{Make every warning contestable.} A distribution can direct attention, but the judgment belongs at the level of text and context. Reviewers can open the records behind an indicator, compare alternative matches, inspect summary claims, and record a revision. Recomputing the audit after an edit reveals whose visibility increased and whose declined. This movement from pattern to source to judgment prevents a single score from standing in for deliberative quality.

\begin{figure}[t]
\centering
\resizebox{\textwidth}{!}{\begin{tikzpicture}[
  font=\small,
  box/.style={draw=ink, line width=0.65pt, rounded corners=3pt, fill=white,
    align=center, inner sep=4pt},
  cell/.style={draw=accent, line width=0.55pt, rounded corners=2pt, fill=white,
    align=center, inner sep=3pt, text width=3.05cm, minimum height=1.15cm},
  term/.style={draw=ink, line width=0.65pt, rounded corners=3pt, fill=surface,
    align=center, inner sep=4pt},
  flow/.style={-{Stealth[length=2.2mm]}, line width=0.8pt, draw=ink},
  audit/.style={-{Stealth[length=2.2mm]}, line width=0.7pt, draw=muted, dashed},
  lab/.style={muted, font=\scriptsize\itshape},
]
\node[box, text width=3.2cm] (nin) at (0,5.0)
  {\textbf{Contributed records}\\[1pt]\scriptsize responses,\\testimony, comments};
\node[box, text width=3.4cm] (nsyn) at (5.1,5.0)
  {\textbf{Declared synthesis}\\[1pt]\scriptsize generated, extractive, human\\[2pt]
   \scriptsize\itshape generation, evaluation and\\[-1pt]\scriptsize\itshape embedding set separately};
\node[box, text width=3.0cm] (nsum) at (10.4,5.0)
  {\textbf{Summary}\\[1pt]\scriptsize account used for review or decision support};
\draw[flow] (nin) -- (nsyn);
\draw[flow] (nsyn) -- (nsum);
\begin{scope}[on background layer]
  \node[draw=muted, dashed, line width=0.6pt, rounded corners=4pt,
    fit=(nin)(nsyn)(nsum), inner sep=9pt] (trans) {};
\end{scope}
\node[lab, anchor=south west, inner sep=1.5pt] at ([xshift=3pt]trans.north west)
  {the source collection, synthesis activity, and resulting artifact};

\node[cell] (c1) at (0,2.4)
  {\textbf{Alignment}\\[1pt]\scriptsize strongest sentence match,\\\scriptsize score distribution};
\node[cell] (c2) at (3.55,2.4)
  {\textbf{Evidence trail}\\[1pt]\scriptsize source $\to$ sentence,\\\scriptsize claim passages};
\node[cell] (c3) at (7.1,2.4)
  {\textbf{Expression review}\\[1pt]\scriptsize language, affect,\\\scriptsize tension annotations};
\node[cell] (c4) at (10.5,2.4)
  {\textbf{Stability}\\[1pt]\scriptsize resampling and\\\scriptsize gated grouping};
\node[lab, align=center] (abstain) at (5.25,1.28)
  {stages that fail, abstain, or do not pass their gates are withheld and remain visible};
\begin{scope}[on background layer]
  \node[draw=accent, line width=0.9pt, fill=panelfill, rounded corners=4pt,
    fit=(c1)(c2)(c3)(c4)(abstain), inner xsep=8pt, inner ysep=12pt] (panel) {};
\end{scope}
\node[accent, font=\footnotesize\bfseries, anchor=north west, inner sep=1.5pt]
  at ([yshift=-1pt,xshift=3pt]panel.north west)
  {PROVENANCE LAB $\cdot$ SOURCE-LEVEL AUDIT};
\draw[audit] (nin.south) -- (nin.south|-panel.north);
\draw[audit] (nsum.south) -- (nsum.south|-panel.north);

\node[box, text width=3.1cm] (rev) at (0,-0.1)
  {\textbf{Human review}\\[1pt]\scriptsize inspect text and record judgment};
\node[box, text width=2.7cm] (revi) at (4.3,-0.1)
  {\textbf{Revise or retain}\\[1pt]\scriptsize preserve an\\explicit decision};
\node[box, text width=2.9cm] (rec) at (8.1,-0.1)
  {\textbf{Recompute}\\[1pt]\scriptsize compare under the run configuration};
\node[term, text width=2.8cm] (out) at (12.0,-0.1)
  {\textbf{Reviewed synthesis}\\[1pt]\scriptsize plus unsigned audit record};
\draw[flow] (rev) -- (revi);
\draw[flow] (revi) -- (rec);
\draw[flow] (rec) -- (out);
\draw[audit] (panel.south -| rev.north) -- (rev.north);
\draw[flow] (rec.north) -- (rec.north|-panel.south);
\end{tikzpicture}}
\caption{\textbf{Provenance Lab workflow.} Contributed records enter a declared synthesis process. The audit connects source records, summary sentences, and claims while keeping missing and unassessed cases visible. Optional model-coded reviews remain separate from deterministic measurements. A reviewer follows a warning to its evidence, retains or revises the summary, recomputes the audit, and exports the reviewed artifact with an unsigned provenance record. Solid arrows show workflow actions. Dashed arrows connect stored material to the audit and audit results to human review.}
\label{fig:overview}
\end{figure}
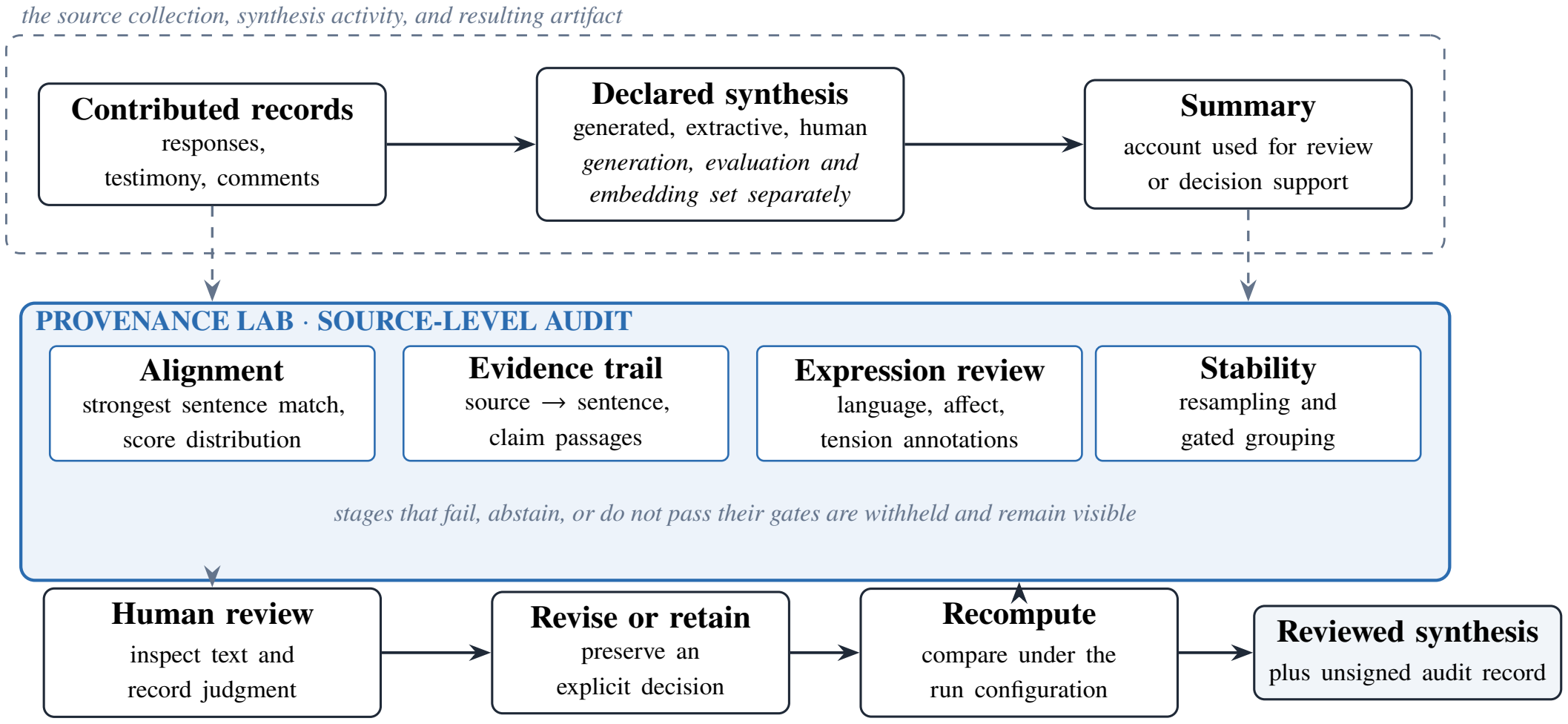

Appendix~\ref{app:checks} describes the software checks behind these safeguards. Figure~\ref{fig:overview} summarizes the workflow. Contributed records and a declared summary enter the audit layer, which produces evidence candidates and bounded indicators. Review sits between those signals and any revised public account. This placement matters in civic settings because the interface can make a possible loss visible while leaving acceptance, rejection, and revision attributable to a reviewer.

\section{Provenance Lab}

\subsection{Preserving and tracing contributions}

Voice survival refers here to whether a contribution's distinctive claim, target, condition, or stance remains traceable in the summary. It is not a binary property inferred from one score. Semantic alignment can locate related wording, while source-unit and claim-level views help a reviewer examine what changed within that relationship. Representational erasure remains a review hypothesis until the linked evidence has been inspected.

The audit begins by preserving contribution boundaries. Each answer remains distinct from the question that prompted it, and short, numeric, uncertain, dissenting, or refusing responses remain in the source census. In multi-question consultations, the prompt is retained as context while the answer remains the evidentiary unit. This matters because preprocessing can remove forms of participation before the summary itself is examined.

The reviewer declares whether the summary was generated during the audit, produced elsewhere, written by a person, developed through mixed authorship, or has unknown origin. This declaration prevents pasted or revised text from acquiring a false production history. Generation, automated evaluation, and embedding remain separate roles, which makes external summaries auditable and makes model self-review visible. Failures and fallbacks remain attached to the run instead of being attributed to the requested model.

Core analysis embeds source records and summary sentences in one verified semantic space. When dense embedding fails, the complete comparison is recomputed with a local lexical method so that vectors from different spaces are never combined. For source record $i$, strongest-sentence alignment is
\begin{equation}
r_i=\max\left(0,\max_j \cos(\mathbf{x}_i,\mathbf{s}_j)\right),
\label{eq:representation}
\end{equation}
where $\mathbf{x}_i$ and $\mathbf{s}_j$ are embeddings from the same run. The measure asks whether at least one summary sentence is semantically close to the source record. A low value identifies a record that may warrant inspection. A high value establishes neither complete preservation nor agreement because topic overlap can remain high when stance, target, or conditions change. Reviewers can use the score to decide which records to read first.

The overview reports valid comparisons, mean alignment, dispersion across records, and a gap between records distant from the source centroid and the remainder. These summaries direct attention to uneven visibility. Their meaning depends on the source collection and carries no independent fairness interpretation. Participant resampling shows sensitivity to reweighting the observed records and is not treated as population inference without a defensible sampling design.

Source maps can reveal computational structure, but they can also invite a reviewer to mistake geometric clusters for natural stakeholder groups. Provenance Lab displays a semantic partition only when it remains stable across repeated fits, subsamples, neighbourhood checks, and plausible methodological choices. If no candidate passes, records remain unpartitioned and analyses that require groups are withheld. A passing partition still requires substantive interpretation and cannot be treated as a demographic category.

\subsection{Examining transformation and apparent consensus}

The interface connects three forms of inquiry. Source-level views ask whose contributions remain visible. Transformation views examine whether specificity, target, stance, urgency, or conditions changed. Claim-level views ask which passages can support assertions in the final synthesis. Table~\ref{tab:views} shows how the workspaces serve these questions and bounds the interpretation of their outputs.

\begin{table}[t]
\caption{Interface workspaces and the conclusions they support.}
\label{tab:views}
\centering
\small
\begin{tabular}{@{}Y{0.25\textwidth}Y{0.31\textwidth}Y{0.34\textwidth}@{}}
\toprule
Workspace & Principal views & Permitted interpretation \\
\midrule
Overview and review priorities & Alignment distribution, quadrants, resampling, declared groups & Descriptive score patterns and candidates for review \\
Participant and topic spaces & Semantic map, mutual-neighbour network, knowledge graph & Source geometry and model-extracted concepts \\
Associations and evidence & Contribution matches, association map, claim evidence, sentence-source view & Retrieval candidates and model-coded support labels \\
Discourse and counterfactual audit & Transformation, affect, tension, rhetoric, leave-one-out (LOO) analysis, strategies & Exploratory expression review and source-space sensitivity \\
\bottomrule
\end{tabular}
\end{table}

The source-unit transformation module begins with every sentence-like answer unit. For each assessed unit, the evaluator returns a category such as preserved, softened, abstracted, target-shifted, or dropped. Mapped units require an exact summary span. A unit is marked dropped when an assessment finds no matching span. Uncertain and unassessed outputs remain separate, and participant-balanced rates prevent long responses from dominating. The categories are automated review prompts whose coverage travels with every rate.

Affective-expression review examines target-relative valence, expressed intensity, urgency, broad emotion families, and stance within the same verified alignment. Its subject is language expressed in the record, rather than a contributor's internal emotional state. The interface reports attenuation, amplification, shifts, omissions, uncertainty, and model-reported certainty without collapsing them into a generic emotion score.

\begin{figure}[p]
\centering
\includegraphics[width=\textwidth]{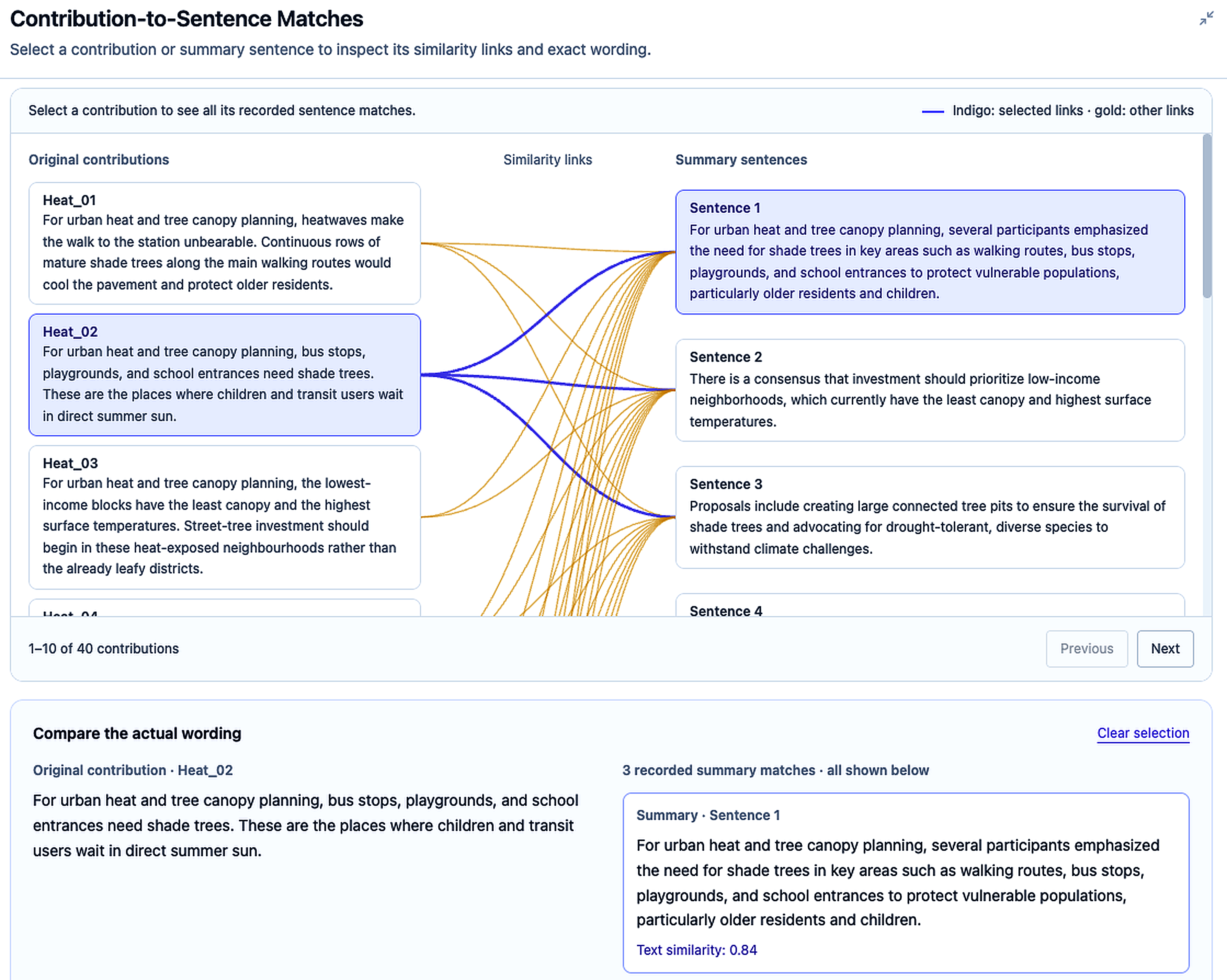}
\caption{\textbf{Tracing a contribution into the summary.} Selecting \texttt{Heat\_02} highlights its three stored sentence matches. Sentence~1 carries over the proposed locations for shade trees. The contribution names children and transit users, while the summary sentence names older residents and children.}
\label{fig:contribution-mapping}
\end{figure}

Figure~\ref{fig:contribution-mapping} shows how a reviewer follows one contribution into the summary. Selecting \texttt{Heat\_02} highlights its stored similarity links and places the original submission alongside the selected summary sentence. The comparison lets the reviewer check what happened to specific details, such as the places and people the contribution identifies.

\begin{figure}[p]
\centering
\includegraphics[width=\textwidth]{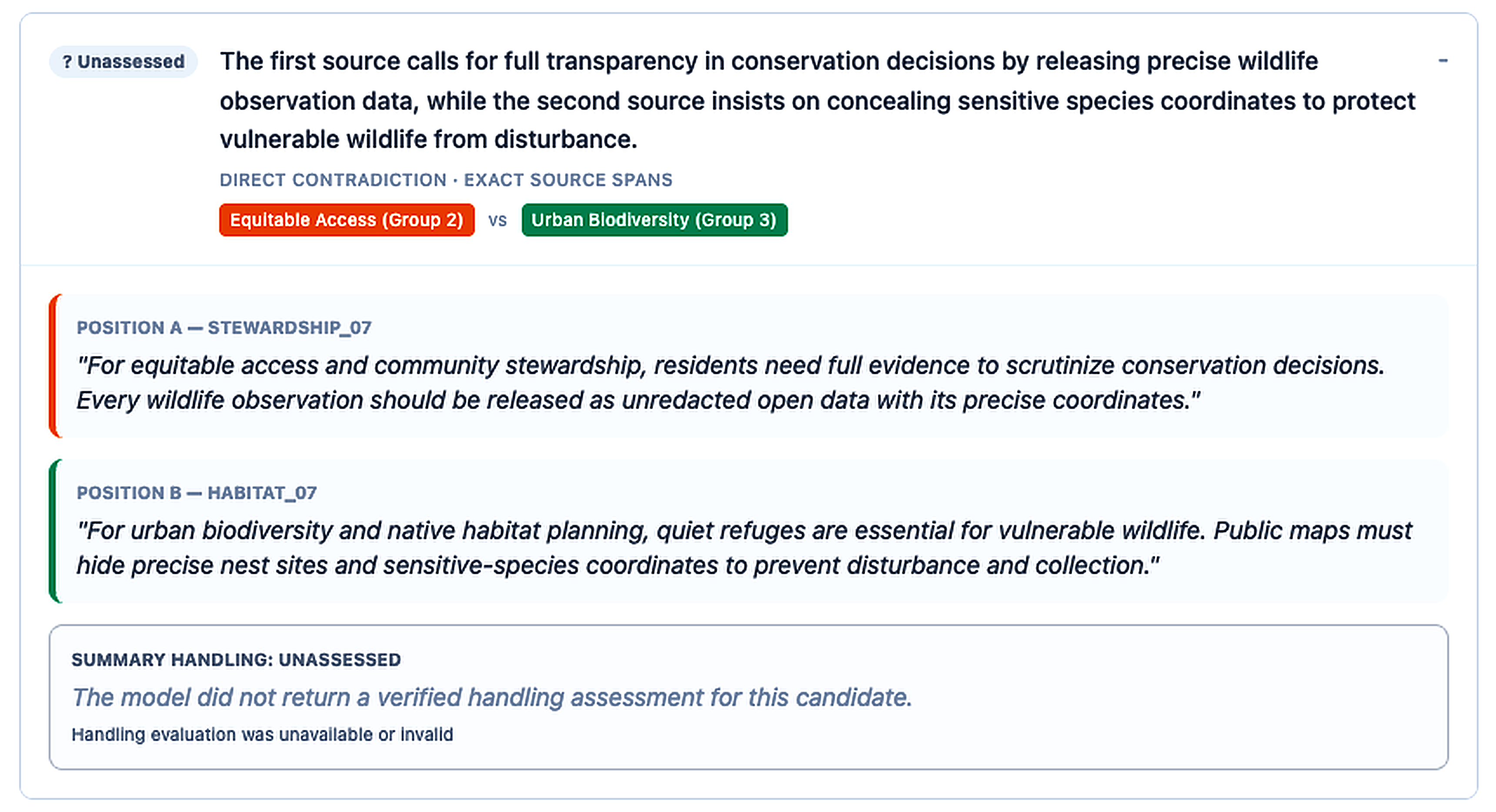}
\caption{\textbf{Inspecting a disagreement in the source record.} \texttt{Stewardship\_07} calls for releasing precise wildlife observation locations so residents can scrutinize conservation decisions. \texttt{Habitat\_07} calls for withholding sensitive locations to protect wildlife. The panel displays both excerpts and the unassessed status of their treatment in the summary.}
\label{fig:tension-mapping}
\end{figure}

Tension Mapping looks for disagreements in sampled pairs of contributions across groups that passed the stability checks. It presents the two source excerpts and a separate assessment of how the summary handled the disagreement. In Figure~\ref{fig:tension-mapping}, the app shows opposing proposals about publishing precise wildlife locations. The handling status is \emph{Unassessed} because the app has no verified assessment of how the summary treated the disagreement.

Claim review identifies assertions in summary sentences and retrieves candidate passages for each. It keeps a sentence whole when splitting it would change its meaning.

A reviewer can edit the summary directly or request a model-assisted draft based on review candidates. Applying an edit launches a fresh audit and shows which records gained or lost alignment. Numeric changes appear with a warning when the underlying semantic spaces differ. Appendix~\ref{app:walkthrough} follows a claim through a saved revision. Appendix~\ref{app:source-comments} illustrates two ways to inspect the original comments, and Appendix~\ref{app:claim-attribution} shows the evidence labels for individual claims.

Participants can receive the audited summary and return structured challenges through an offline exchange. Integrity checks bind a response to the expected summary, run, and participant roster. The mechanism supports version-specific contestation, while personal authentication and a secure institutional return channel remain outside the application.

The final export preserves the connection among the source set, declared summary origin, automated assessments, revisions, and stated limitations using W3C PROV concepts \citep{moreau2013}. It supports inspection and change detection but remains unsigned. Consequential deployment therefore requires institutional provisions for identity, consent, access, retention, withdrawal, and key custody.

\section{Empirical assessment of evidence retrieval}

The empirical assessment focuses on evidence retrieval because this mechanism connects an aggregate warning to the source text a reviewer will inspect. Poor ranking can obscure the contribution that warrants attention and direct review toward unrelated material. Retrieval quality is therefore a necessary, though incomplete, test of whether the evidence-linked workflow functions as intended.

\begin{figure}[t]
\centering
\includegraphics[width=\textwidth]{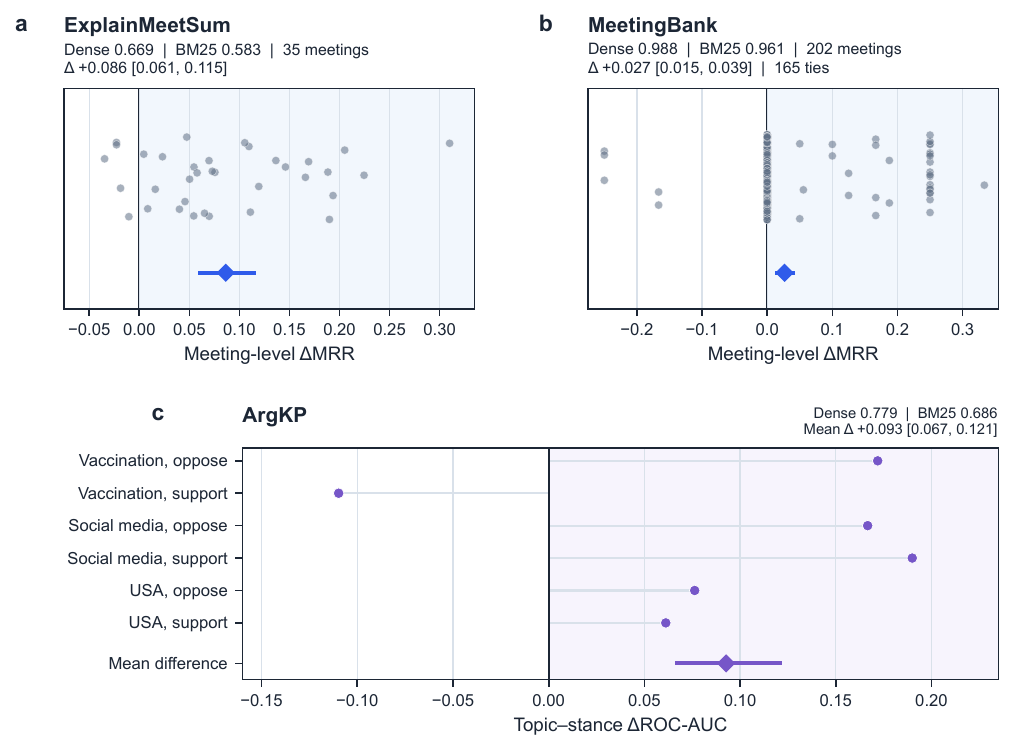}
\caption{\textbf{Semantic search (Dense) versus keyword search (BM25).} In \textbf{a} (ExplainMeetSum) and \textbf{b} (MeetingBank), each dot shows the difference in mean reciprocal rank for one meeting. Dots are spread vertically for visibility. In MeetingBank, 165 of 202 meetings tied. In \textbf{c} (ArgKP), each dot shows the difference in ROC-AUC for one topic and stance, comparing how well each method distinguishes matching from non-matching argument--key-point pairs. Shading marks the side that favours semantic search. Diamonds mark the mean differences, with horizontal lines showing 95\% bootstrap intervals.}
\label{fig:eval-retrieval}
\end{figure}

Three released datasets provide independent evidence links for the primary retrieval comparison: human-aligned meeting evidence, official civic minutes, and human-labelled argument-to-key-point pairs. The comparison holds the candidate text fixed and varies only the scoring function. Semantic search (labelled Dense in Figure~\ref{fig:eval-retrieval} and Table~\ref{tab:eval}) uses the \href{https://ollama.com/library/qwen3-embedding:8b}{\texttt{qwen3-embedding:8b}} model in Ollama to embed each query and passage and rank passages by similarity. The keyword search method, BM25, is a standard ranker that rewards shared informative terms while adjusting for term frequency and passage length \citep{robertson2009}. It is used with $k_1=1.2$ and $b=0.75$ as a conventional lexical reference point and was not tuned for each dataset. A tuned lexical system could narrow the reported gaps. Two standard measures are reported. Mean reciprocal rank (MRR) averages the score for each query. The first correct passage earns a score of 1 if it ranks first, 1/2 if it ranks second, and so on. ROC-AUC is the probability that a matching pair outscores a non-matching pair, with half credit for ties. A score of 0.5 is chance-level ranking. Two further analyses compare alignment scores with expert summary ratings and test whether the scores help predict participant preferences.

\begin{figure}[t]
\centering
\includegraphics[width=\textwidth]{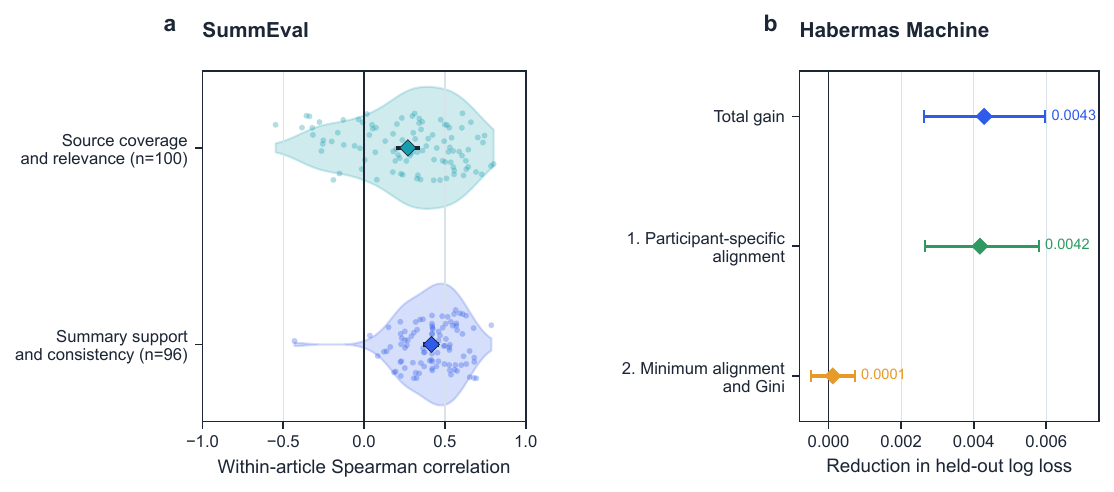}
\caption{\textbf{Agreement with expert ratings and prediction of participant preferences.} \textbf{a}, Each dot shows how closely the automated scores and experts agreed when ranking summaries of one news article. The two rows show 100 relevance correlations and 96 consistency correlations. Expert consistency scores were constant for the other four articles. Violins show the distribution. Diamonds and lines show mean correlations and 95\% intervals from resampling articles. \textbf{b}, The Habermas Machine results show reductions in held-out log loss for participant preference predictions. Values to the right of zero indicate better predictions. The top row shows the total gain over the baseline. The next two rows show the gain from adding participant-specific alignment and then from adding minimum alignment and Gini. Lines show 95\% intervals from resampling whole deliberation sessions.}
\label{fig:eval-convergent}
\end{figure}

The evaluation protocol set out four predictions before the final tests. Three asked whether semantic search would rank evidence higher than keyword search (Table~\ref{tab:eval} and Figure~\ref{fig:eval-retrieval}). The fourth asked whether alignment scores would help predict participant preferences (Figure~\ref{fig:eval-convergent}). Other comparisons are descriptive. To estimate uncertainty, the analysis resampled whole meetings, articles, or deliberation sessions so that related observations stayed together. For ArgKP, it resampled arguments within each topic and stance.

\begin{table}[!ht]
\caption{Evidence retrieval on three benchmarks. MRR differences are averaged across meetings, and ROC-AUC differences across topic-and-stance groups. Intervals use 2,000 resamples of meetings or arguments.}
\label{tab:eval}
\centering
\small
\begin{tabular}{@{}Y{0.21\textwidth}Y{0.32\textwidth}Y{0.37\textwidth}@{}}
\toprule
Dataset & Reference signal & Primary result \\
\midrule
ExplainMeetSum \citep{kim2023} & Human-aligned central meeting evidence & Mean MRR difference (Dense $-$ BM25) $+0.0863$, 95\% CI $[0.0607,0.1147]$ \\
MeetingBank \citep{hu2023} & Official minute-transcript links & Mean MRR difference (Dense $-$ BM25) $+0.0267$, 95\% CI $[0.0152,0.0393]$ \\
ArgKP \citep{barhaim2020,friedman2021} & Released human-labelled argument-key-point pairs & Mean ROC-AUC difference (Dense $-$ BM25) $+0.0927$, 95\% CI $[0.0669,0.1207]$ \\
\bottomrule
\end{tabular}
\end{table}

{\clubpenalties 3 10000 10000 0
ExplainMeetSum contained 35 meetings and 1,525 eligible queries. The mean reciprocal rank across meetings was 0.669 for semantic search and 0.583 for BM25 (Figure~\ref{fig:eval-retrieval}a). The first result from semantic search was a sentence that annotators had marked as central evidence 55.9\% of the time, on average across meetings. Provenance Lab therefore shows several candidates so reviewers can look beyond the first match. In MeetingBank, the two methods tied in 165 of 202 meetings. The search had just two candidate passages in 136 meetings, including 117 of the ties. ArgKP \citep{barhaim2020,friedman2021} included 3,426 labelled pairs across six topic-and-stance groups. The analysis used only the released labels. Semantic search performed better in five groups. BM25 performed better in the group labelled \emph{Vaccination, support} (Figure~\ref{fig:eval-retrieval}c).\par}

SummEval provides expert ratings for 1,600 summaries of 100 news articles \citep{fabbri2021}. Experts rated relevance by judging whether a summary included the article's important information. They rated consistency by judging whether the article supported the facts in the summary. For each article, the analysis compared how the experts ranked its summaries with how two similarity scores ranked them. Source coverage matches each article sentence to its closest summary sentence. Its average rank correlation (Spearman) with relevance was 0.271. Summary support matches each summary sentence to its closest article sentence. Its average correlation with factual consistency was 0.417 across 96 articles. For each of the other four articles, all 16 summaries had the same expert consistency score, so the correlation was undefined. Higher scores tended to go with higher expert ratings, but the relationship varied across articles (Figure~\ref{fig:eval-convergent}).

{\interlinepenalty=10000
The Habermas Machine release records how participants ranked candidate group statements \citep{tessler2024}. The baseline used statement length, display position, and average alignment with the group to predict which statement a participant would prefer in a pair. Each of five cross-validation folds held out whole deliberation sessions. Adding both sets of audit features reduced held-out log loss from 0.63376 to 0.62947, a gain of 0.00429 (95\% CI $[0.00262,0.00597]$). Most of the gain came from adding participant-specific alignment, which measures how closely the statement matched the ranking participant's own contribution. This step reduced log loss by 0.00417 (95\% CI $[0.00266,0.00579]$). Adding minimum alignment and the Gini coefficient reduced log loss by a further 0.00012 (95\% CI $[-0.00048,0.00072]$). The interval includes zero, so the second step added no detectable predictive value.\par}

The Shapley calculation estimates how much each input contributes to a score when combined with different subsets of the other inputs \citep{shapley1953}. In 12 test cases, the score for every possible subset was specified in advance. The software returned the same value for each input as an independent exact calculation. For the Monte Carlo approximation, root mean squared error averaged 0.000716 and did not exceed 0.002799.

Across the three retrieval benchmarks, the fixed \texttt{qwen3-embedding:8b} configuration improved the prespecified aggregate ranking measures over the fixed BM25 baseline. The benchmarks leave open how other embedding models would rank the same passages. Agreement with expert ratings was moderate, and the association with participant preferences was small.

\section{Discussion}

Civic synthesis exercises representational authority because the resulting text may become the version of participation that decision makers and the public actually read. Erasure need not take the form of complete deletion. A summary can preserve a topic while removing who is affected, why a position was taken, the conditions attached to support, or the fact that participants disagreed. When these differences are absorbed into a broad statement of agreement, the synthesis may appear more settled than its source record.

Provenance Lab makes such transformations available for contestation. It brings aggregate patterns, source wording, claim evidence, revision, and participant challenge into one review workflow. These outputs function as prompts for inspection. They cannot determine whether a contribution was fairly treated or whether consensus exists. Their value lies in making the basis of a collective statement easier to question before it hardens into an official account. Table~\ref{tab:status} states, for each major output, the use it currently supports and the boundary that applies to it.

\begin{table}[!ht]
\caption{Scientific status assigned to major outputs in the current application.}
\label{tab:status}
\centering
\small
\begin{tabular}{@{}Y{0.24\textwidth}Y{0.31\textwidth}Y{0.35\textwidth}@{}}
\toprule
Output & Immediate use & Required boundary \\
\midrule
Strongest-sentence alignment & Retrieve a plausible match for each source & Similarity, without entailment, completeness, authorship, or causality\\
\addlinespace[2pt]
Gini and distinct-record gap & Describe score inequality and records distant from the source centroid & Descriptive geometry, without an independent fairness interpretation\\
\addlinespace[2pt]
Semantic partition & Organize source-space exploration after stability gates & Computational grouping, without demographic or natural-category meaning\\
\addlinespace[2pt]
Transformation and affect & Review exact-span model-coded language changes & Exploratory automated coding with coverage and abstention\\
\addlinespace[2pt]
Claim evidence & Inspect retrieved passages and support labels & Candidate evidence, without a verified citation guarantee\\
\addlinespace[2pt]
Counterfactual utility attribution & Estimate contribution to a specified generated-summary utility & Stochastic utility attribution, without historical causal identification\\
\addlinespace[2pt]
PROV-aligned record & Preserve execution facts, digests, assessments, and limitations & Unsigned change-detection record, without identity authentication\\
\bottomrule
\end{tabular}
\end{table}

The separation between execution provenance and semantic association supports two different forms of accountability. Production records can show which system created or revised an artifact. Semantic links can help a reviewer locate related text. Neither alone establishes whose judgment shaped the final account. Keeping them separate reduces the risk that interface confidence will stand in for institutional evidence.

The system also exposes where application-level safeguards end. It begins with the records supplied to it. It cannot discover people who were never invited, recruitment failures, inaccessible participation channels, or institutional pressure that shaped the source collection. Declared groups may be incomplete or inappropriate. A stable semantic cluster is not a demographic group. A participant-response package can detect mismatched content without authenticating the respondent. The audit record can detect modification without proving authorship. These gaps require institutional controls and, in many settings, legal and participatory governance.

The empirical evaluation covers evidence retrieval and selected alignment measures. The reliability of the automated transformation, affect, tension, rhetoric, and knowledge-graph outputs still needs to be assessed against qualified human judgments. The evaluation is English-dominant. Multilingual evaluation should test retrieval across languages and involve reviewers familiar with how stance, disagreement, and conditions are expressed in the communities being studied.

Whether Provenance Lab helps facilitators find more omissions or produce better revisions remains an open question. A study could have facilitators review comparable summaries with and without the interface. Independent reviewers could assess which omissions were found and whether edits preserved the relevant positions and conditions. Time spent finding evidence and disagreement among reviewers would help explain any differences in performance.

In a civic consultation, the right to be heard has a practical implication for AI summarization. Participants should be able to see whether their reasons, reservations, and conditions are reflected in the account that reaches decision makers and raise concerns about omissions or changes in meaning. Provenance Lab helps reviewers trace claims back to the original submissions, record challenges, and compare revisions before the summary becomes the official account. Institutions remain responsible for making that review accessible and responding to challenges.

\section{Conclusion}

AI-mediated civic summaries can make a plural record appear more settled than its underlying contributions. Provenance Lab provides an evidence-linked workflow for examining this transformation. It connects summary sentences and claims to candidate source material, keeps missing and failed assessments visible, and supports revision followed by renewed analysis. Semantic search helped locate candidate evidence in the benchmarks. In the annotated meeting benchmark, the first result often missed that evidence. Agreement with expert ratings also varied across articles. The application does not certify fairness or consensus. Its contribution is a practical way to ask, with visible evidence, whose claims remain legible, which differences were weakened, and where a collective statement requires further scrutiny.

\section*{Data and code availability}

Provenance Lab is openly available and includes a \href{https://sachit27.github.io/apps/provenance-lab/}{browser-based demo for educational use}. The demo uses an example consultation and runs without an API key. The \href{https://github.com/sachit27/co-creation-provenance-lab}{GitHub repository} provides the source code under the GNU Affero General Public License v3.0, the saved demo data, regression tests, and \href{https://github.com/sachit27/co-creation-provenance-lab/blob/main/docs/methodology.md}{methodology documentation}. Section~4 identifies the released datasets used for evaluation. The interface screenshots come from the public demo and the original demonstration run.

\begin{ack}
This work received no specific funding. The author declares no competing interests.
\end{ack}

\providecommand{\doi}[1]{\href{https://doi.org/#1}{\textup{DOI}}}
\bibliographystyle{plainnat}
\bibliography{references}

\clearpage
\appendix
\section{Reproducibility and disclosures}
\label{app:checks}
\paragraph{Software checks.}
Tests check that retrieved excerpts retain decisive source wording, unassessed records are kept distinct from omissions, and unstable groups are withheld. They also check that the tool rejects responses tied to a different summary and detects changes to an exported record. Assessing the reliability of automated labels in use would require independent human judgments.

\paragraph{Use of generative AI tools.}
Generative AI tools assisted with code auditing, literature retrieval, interface testing, and language editing. They did not create the external dataset labels or reported numerical results. The author selected the design, inspected the implementation and outputs, verified sources and calculations, and accepts responsibility for the paper.

\section{Walkthrough of the public demo}
\label{app:walkthrough}

The demo uses an example consultation with 40 fictional comments on trees, flooding, biodiversity, and community stewardship. All 40 received a similarity score and none fell below the lower review marker. A reviewer can still find claims that need a closer look. In Claim Evidence, a proposal from \texttt{Heat\_03} supports prioritizing heat-exposed neighbourhoods. The broader claim that participants reached consensus is marked unassessed.

\begin{figure}[!ht]
\centering
\includegraphics[width=\textwidth]{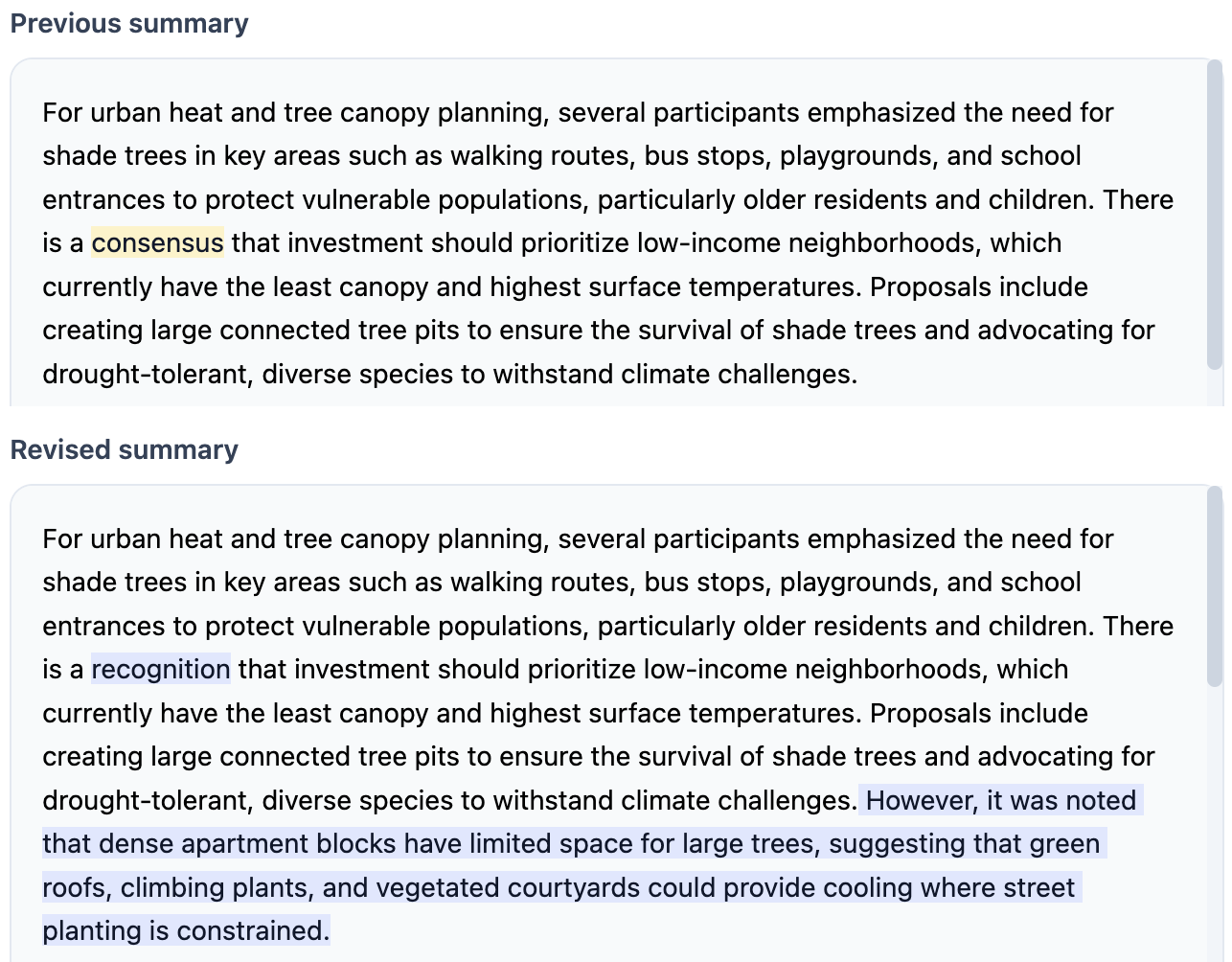}
\caption{\textbf{Reviewing a saved revision in Provenance Lab.} The excerpts show the previous and revised wording. The full revision grew from 256 to 433 words and was scored again on the same 40 comments. The change in length limits what can be inferred about individual edits from the score changes.}
\label{fig:demo-revision}
\end{figure}

The saved revision changes \emph{consensus} to \emph{recognition} and adds a different comment's point about apartment blocks with little space for large trees. Figure~\ref{fig:demo-revision} shows these edits. A reviewer can still ask whose recognition is documented and whether the revised wording should be more specific.

\clearpage
\section{Exploring the source comments}
\label{app:source-comments}

\begin{figure}[!ht]
\centering
\includegraphics[width=\textwidth]{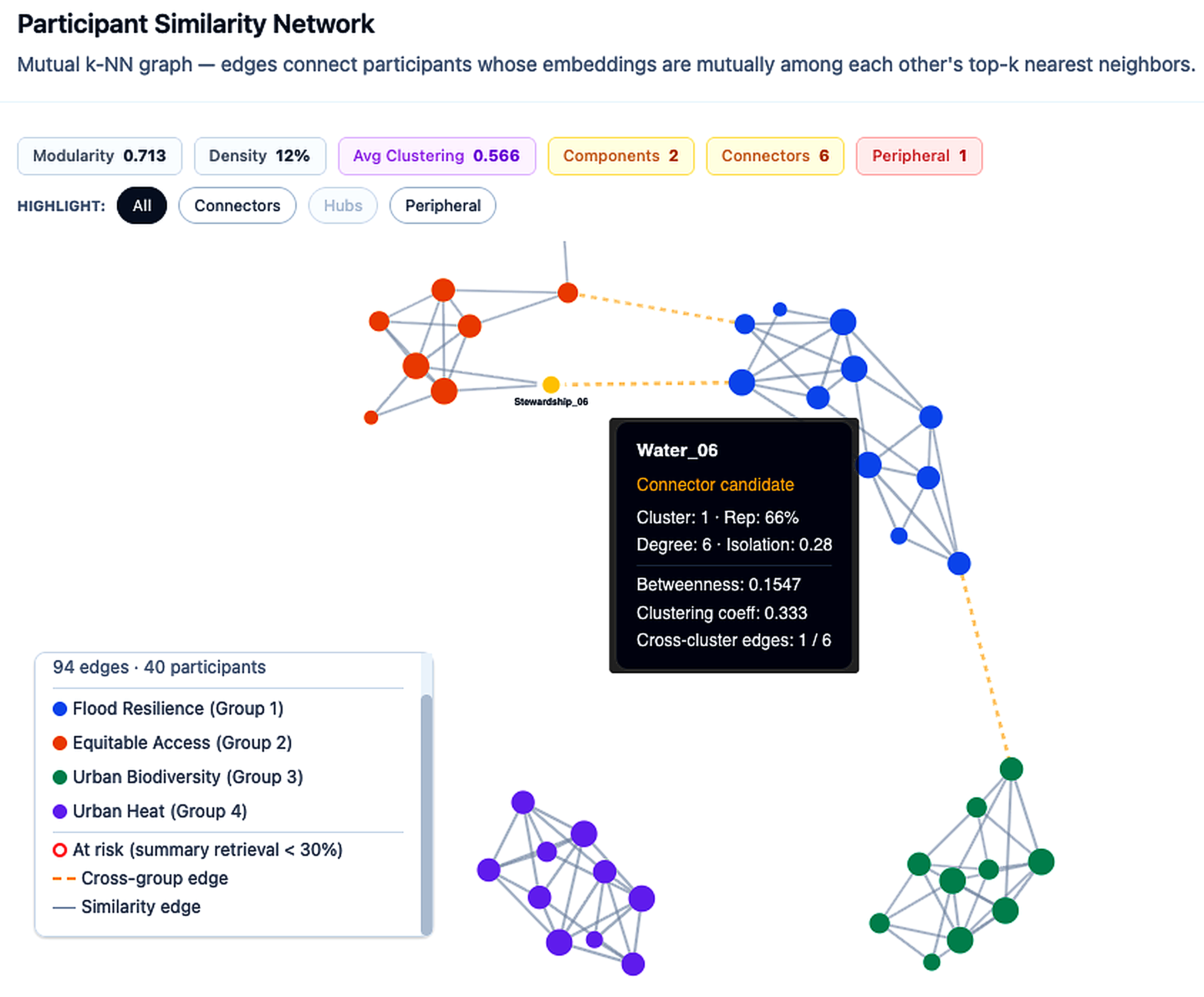}
\caption{\textbf{Participant similarity network.} The dashed orange line connects \texttt{Stewardship\_06}, highlighted in yellow, to \texttt{Water\_06}, shown in the pop-up. Both comments discuss the same community garden. \texttt{Stewardship\_06} asks to keep the plots because families rely on them for food and social support. \texttt{Water\_06} proposes removing every plot to create a stormwater retention area. The text-similarity link helps reviewers find and compare these opposing proposals.}
\label{fig:demo-network}
\end{figure}

In Figure~\ref{fig:demo-network}, each dot is a comment. A line appears when each comment is among the other's six closest matches. The colours show groups computed from text similarity.

\clearpage
\subsection{Leave-one-out input sensitivity}

\begin{figure}[!ht]
\centering
\includegraphics[width=0.95\textwidth,trim=0 5pt 0 10pt,clip]{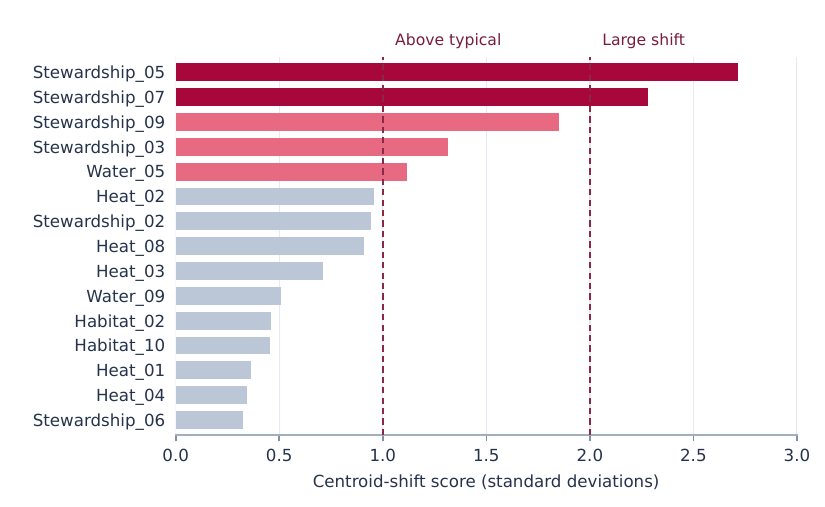}
\caption{\textbf{Leave-one-out input sensitivity.} The chart shows the 15 largest scores in the 40-comment demo. Dashed lines mark one and two standard deviations above the mean shift. The summary stays fixed, so the chart measures sensitivity of the input collection.}
\label{fig:loo-sensitivity}
\end{figure}

Figure~\ref{fig:loo-sensitivity} shows how much the average of the comment vectors shifts when each comment is removed in turn. Longer bars indicate larger shifts relative to the other removals.

\clearpage
\section{Claim attribution in the original demonstration}
\label{app:claim-attribution}

\begin{figure}[!ht]
\centering
\includegraphics[width=\textwidth]{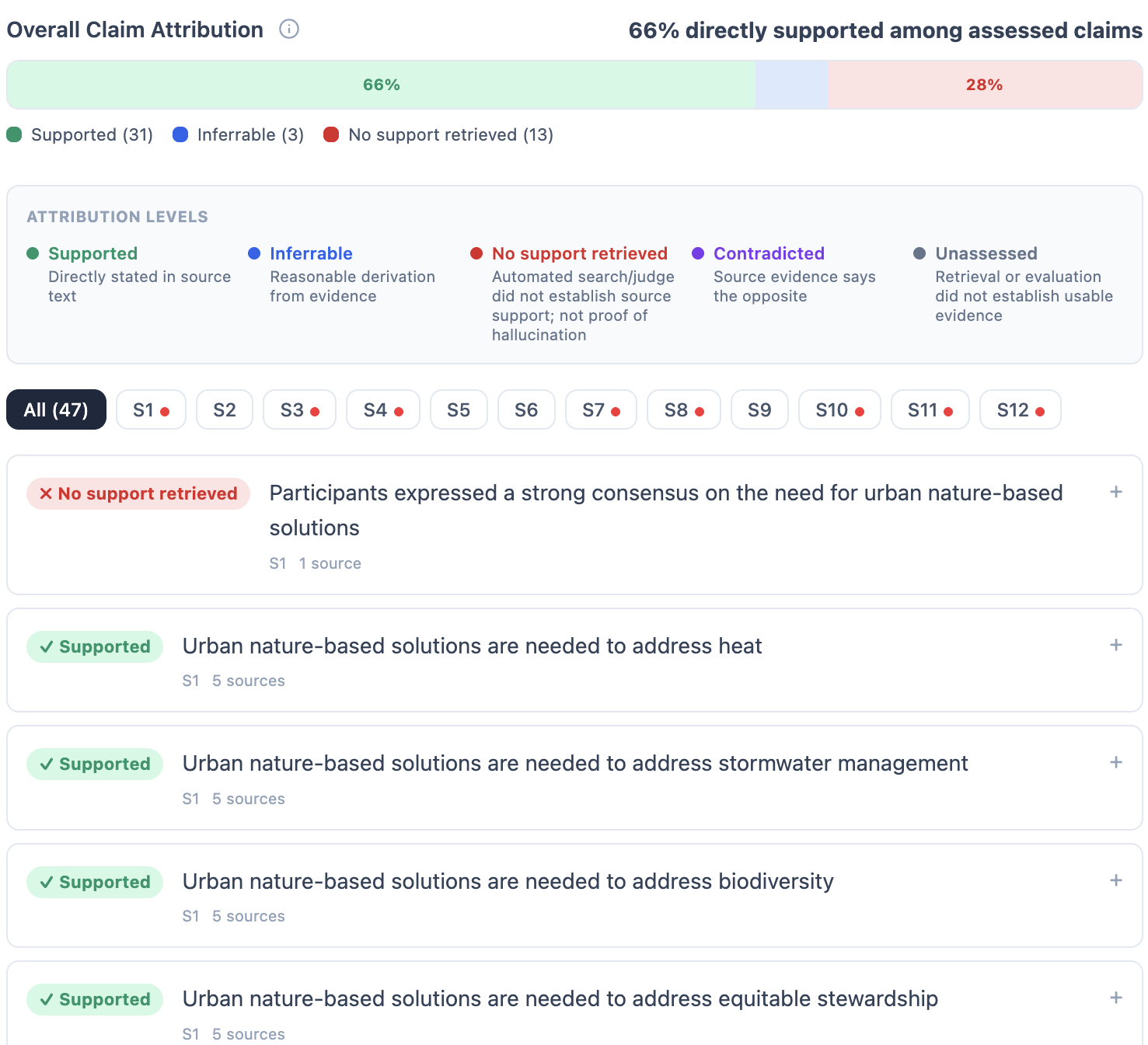}
\caption{\textbf{Overall Claim Attribution in the original demonstration.} Specific proposed actions have candidate support. The broader assertion of strong consensus is marked \emph{No support retrieved}.}
\label{fig:claim-attribution}
\end{figure}

The claim review view (Figure~\ref{fig:claim-attribution}) lists the assertions extracted from the summary and the automated support labels. The overall percentage describes the labels among assessed claims in this demonstration run. Reviewers can open individual claims and inspect the retrieved passages alongside the summary wording.

\end{document}